\documentclass{article}

\usepackage{microtype}
\usepackage{graphicx}
\usepackage{subcaption}
\usepackage{booktabs} 
\usepackage{multirow}

\usepackage{hyperref}

\usepackage[preprint]{icml2026}

\usepackage{amsmath}
\usepackage{amssymb}
\usepackage{mathtools}
\usepackage{amsthm}
\usepackage{mathrsfs}

\usepackage[capitalize,noabbrev]{cleveref}

\theoremstyle{plain}

\theoremstyle{definition}

\theoremstyle{remark}

\usepackage[textsize=tiny]{todonotes}

\icmltitlerunning{It Takes a Good Model to Train a Good Model: Generalized Gaussian Priors for Optimized LLMs}

\begin{document}

\twocolumn[
  \icmltitle{It Takes a Good Model to Train a Good Model:\\ Generalized Gaussian Priors for Optimized LLMs}



  \icmlsetsymbol{equal}{*}

  \begin{icmlauthorlist}
    \icmlauthor{Jun Wu}{tsinghua}
    \icmlauthor{Patrick Huang}{illinois}
    \icmlauthor{Jiangtao Wen}{nyu}
    \icmlauthor{Yuxing Han}{tsinghua}
  \end{icmlauthorlist}

  \icmlaffiliation{tsinghua}{Shenzhen International Graduate School, Tsinghua University}
  \icmlaffiliation{illinois}{Electrical and Computer Engineering, University of Illinois at Urbana-Champaign}
  \icmlaffiliation{nyu}{Computer Science, New York University Shanghai}

  \icmlcorrespondingauthor{Yuxing Han}{yuxinghan@sz.tsinghua.edu.cn}
  \icmlcorrespondingauthor{Jiangtao Wen}{jw9263@nyu.edu}

  \icmlkeywords{Machine Learning, ICML}

  \vskip 0.3in
]



\printAffiliationsAndNotice{}  

\begin{abstract}
Despite rapid progress in large language models (LLMs), the statistical structure of their weights, activations, and gradients—and its implications for initialization, training dynamics, and efficiency—remains largely unexplored. We empirically show that these quantities in LLMs are well modeled by generalized Gaussian (GG) distributions, and introduce a unified, end-to-end optimization framework grounded in this observation. Our contributions are threefold: (1) a GG-based initialization that aligns with trained model statistics, accelerating convergence and improving accuracy; (2) ACT, a progressive activation-constrained training method that reduces redundancy and propagation overhead; and (3) GCT, a gradient-constrained training algorithm that substantially lowers communication cost in distributed training. Experiments across diverse architectures demonstrate consistently smaller, faster models with minimal communication overhead that match or surpass standard baselines. By anchoring LLM optimization in principled statistical modeling, this work advances efficient, scalable, and hardware-aware AI systems.
\end{abstract}

\section{Introduction}

Large language models (LLMs) achieve impressive performance across tasks such as code generation, question answering, and reasoning, but their scale imposes severe challenges for training, deployment, and real-time inference—especially under memory, communication, and power constraints. Existing efficiency methods, including pruning, quantization, and distillation, are typically applied post hoc and focus on data-level redundancy, largely overlooking the underlying statistical structure of model components. A fundamental question, therefore, remains underexplored: what statistical distributions govern weights, activations, and gradients in LLMs, and how can this structure be exploited to train efficient models from the outset?

Recently, BackSlash \citep{pmlr-v267-wu25ai} demonstrated that model parameters are well modeled by a Generalized Gaussian Distribution (GGD) and proposed optimizing distribution entropy during training to enhance compressibility. Unlike post-training precision control methods, BackSlash improves the effectiveness of quantization, sparsification, and entropy coding with negligible performance loss.

Building on this insight, we initialize model weights directly from low-entropy GGDs, promoting inherently low-bitrate representations after convergence. We further show that activations and gradients exhibit even stronger GGD characteristics than parameters. Guided by this observation, we extend distribution-constrained optimization to the activation and gradient domains and introduce tailored constrained training algorithms that enable LLMs to produce naturally more compressible representations throughout training.

Our main contributions are summarized as follows:
\begin{itemize}
    \item \textbf{Generalized Gaussian Structure of LLMs.} We demonstrate that weights, activations, and gradients in LLMs consistently follow generalized Gaussian distributions, with shape parameters typically below 2, especially for activations and gradients.
    \item \textbf{GG-Guided Initialization and Constrained Training.} We propose a GGD-aligned initialization scheme and extend distribution-constrained optimization to activations and gradients, enabling faster convergence, improved generalization, reduced outliers, and rate-stable gradients.
    \item \textbf{End-to-End Framework for Efficient LLMs.} We introduce a unified GG-based framework spanning initialization, training, and inference, yielding smaller, faster models with reduced communication overhead while matching or exceeding standard baselines across architectures and benchmarks.
\end{itemize}

\begin{figure*}[ht]
\centering
\begin{subfigure}[b]{0.33\textwidth}
    \centering
    \includegraphics[width=\linewidth]{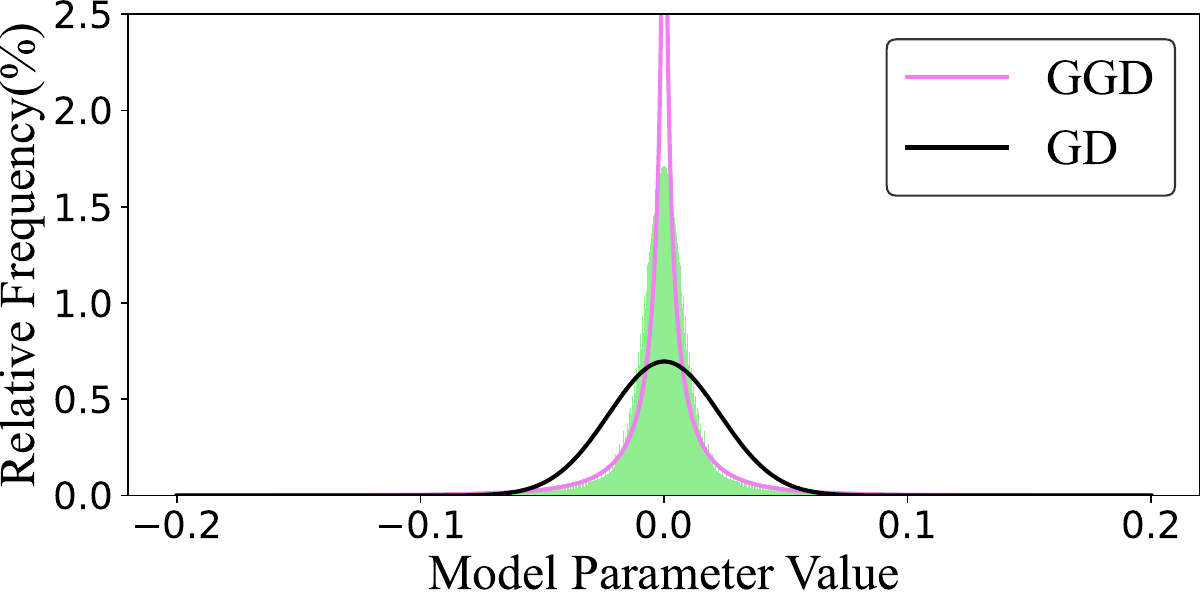}
    \caption{Gemma Weights}
    \label{fig:weight_A}
\end{subfigure}
\begin{subfigure}[b]{0.33\textwidth}
    \centering
    \includegraphics[width=\linewidth]{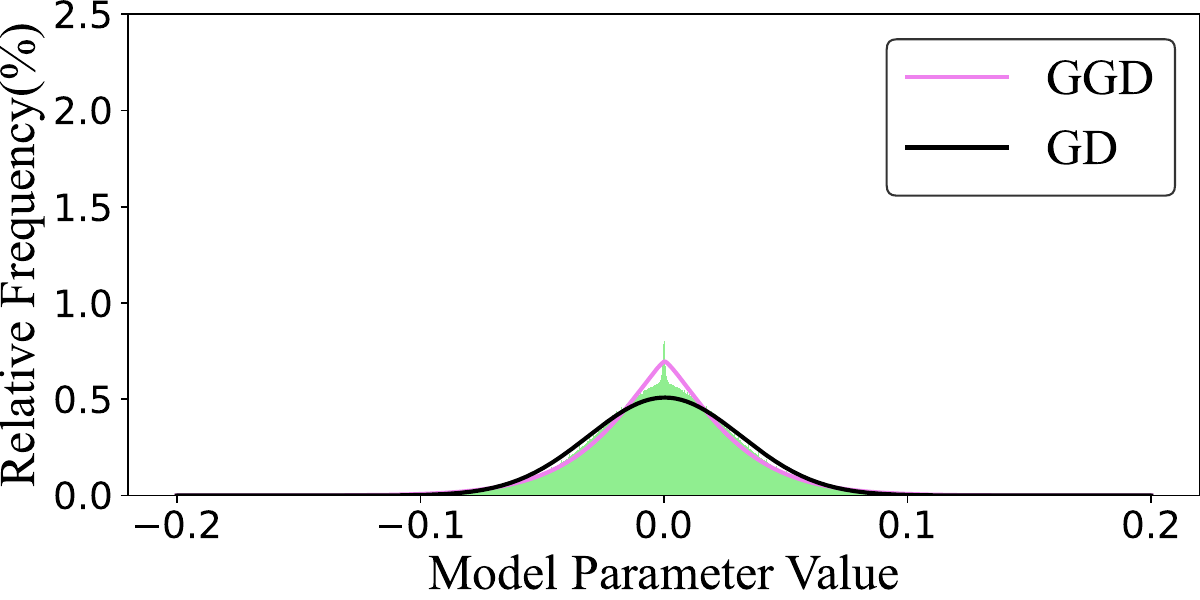}
    \caption{Phi Weights.}
    \label{fig:act_A}
\end{subfigure}
\begin{subfigure}[b]{0.33\textwidth}
    \centering
    \includegraphics[width=\linewidth]{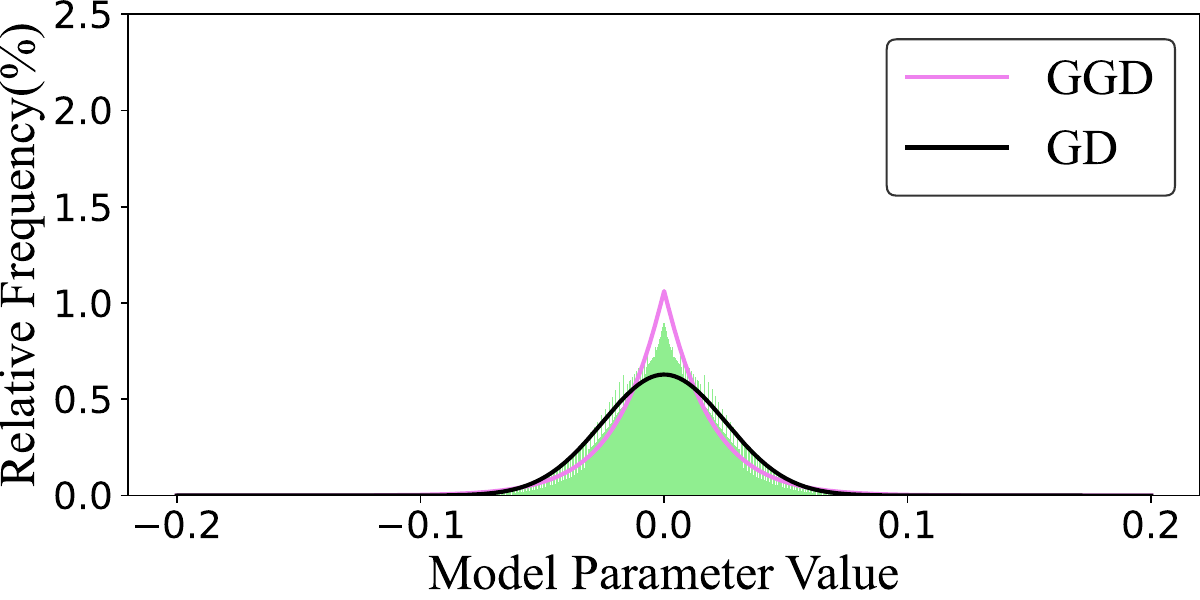}
    \caption{Qwen Weights.}
    \label{fig:grad_A}
\end{subfigure}


\begin{subfigure}[b]{0.33\textwidth}
    \centering
    \includegraphics[width=\linewidth]{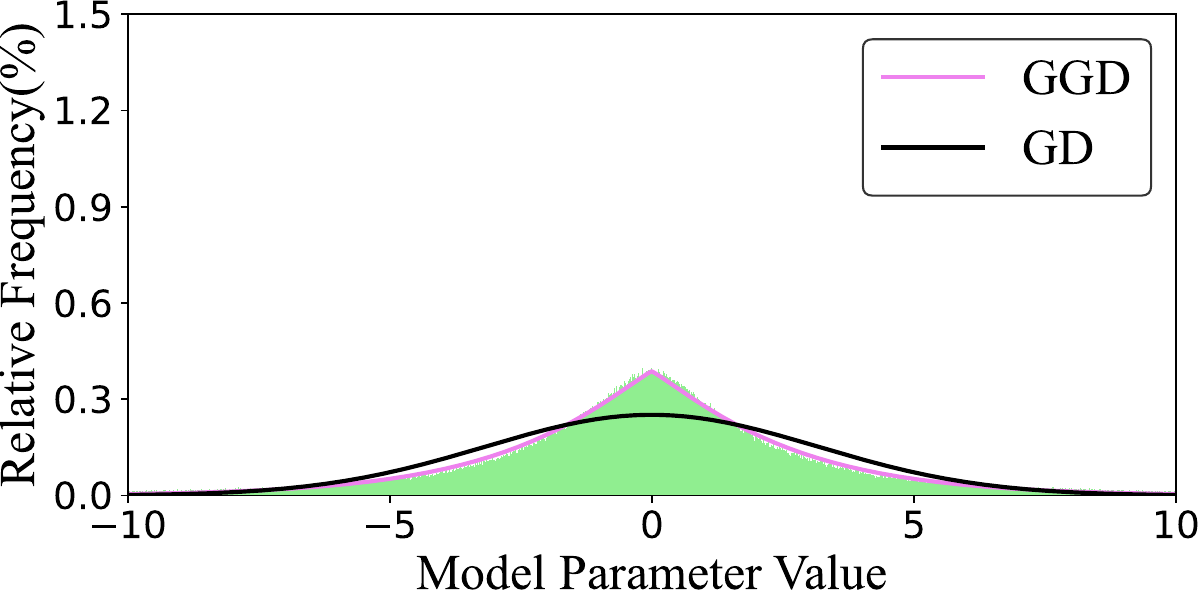}
    \caption{Gemma Activations.}
\end{subfigure}
\begin{subfigure}[b]{0.33\textwidth}
    \centering
    \includegraphics[width=\linewidth]{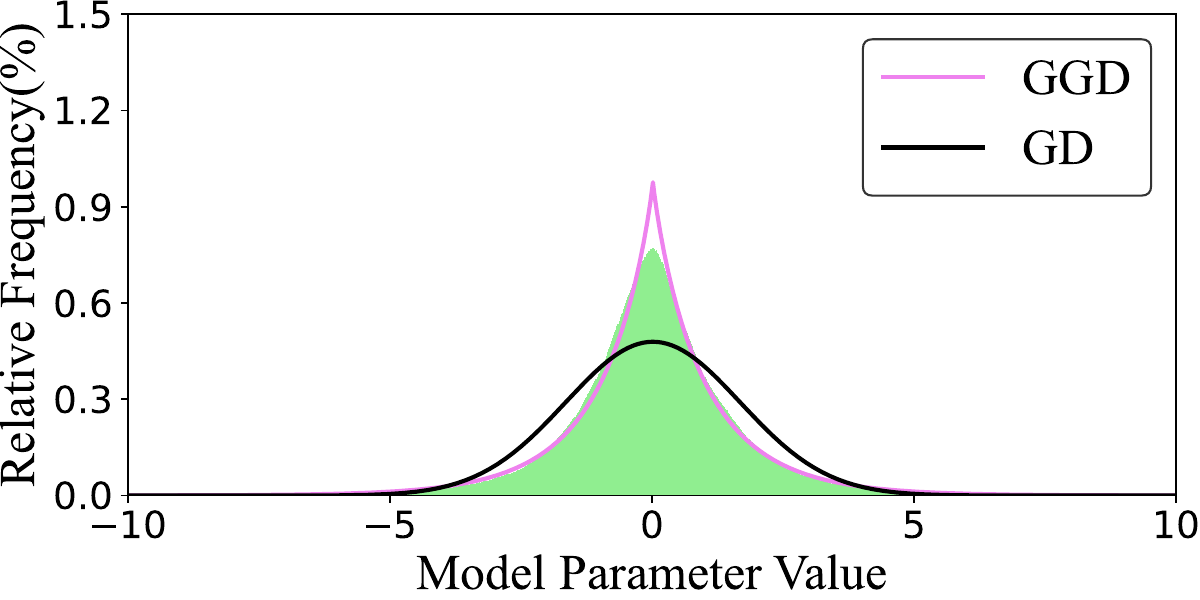}
    \caption{Phi Activations.}
\end{subfigure}
\begin{subfigure}[b]{0.33\textwidth}
    \centering
    \includegraphics[width=\linewidth]{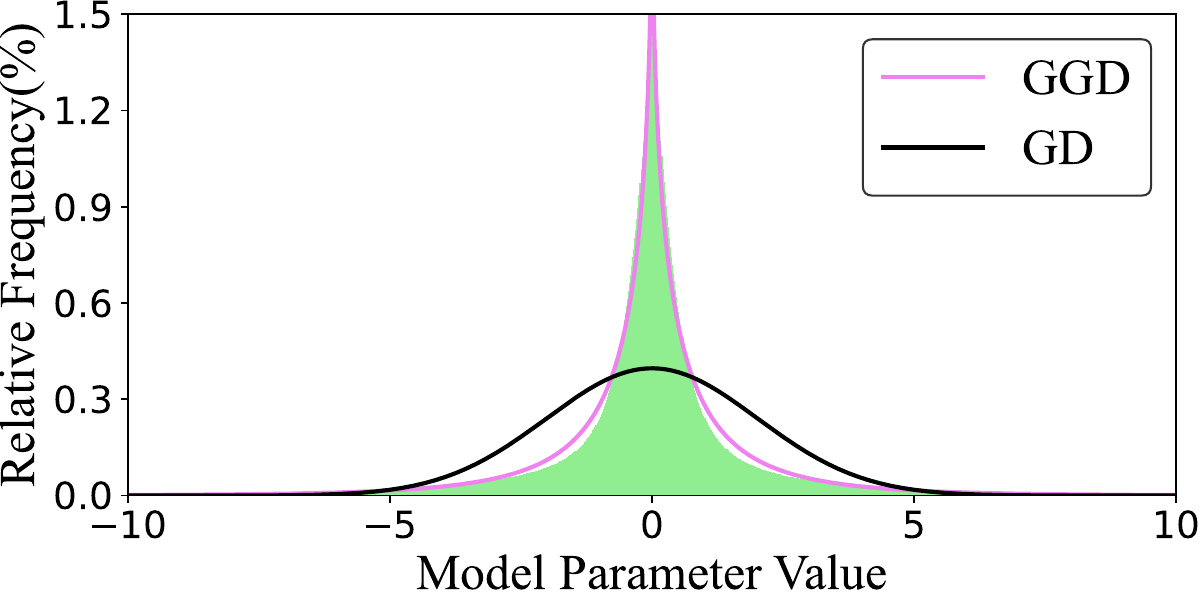}
    \caption{Qwen Activations.}
\end{subfigure}


\begin{subfigure}[b]{0.33\textwidth}
    \centering
    \includegraphics[width=\linewidth]{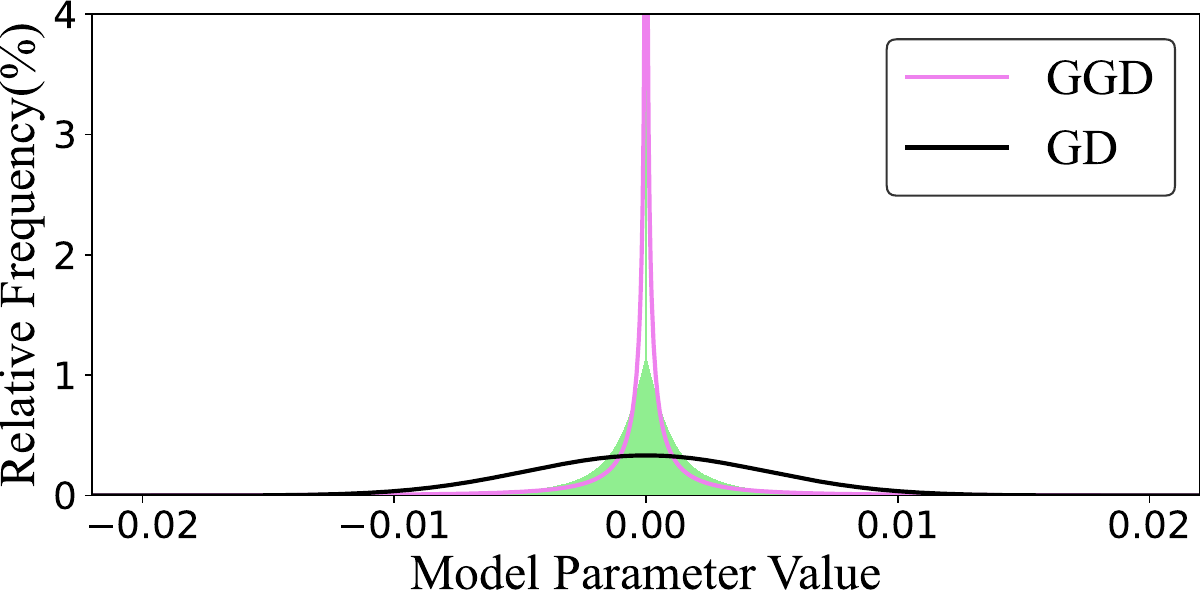}
    \caption{Gemma Gradients.}
\end{subfigure}
\begin{subfigure}[b]{0.33\textwidth}
    \centering
    \includegraphics[width=\linewidth]{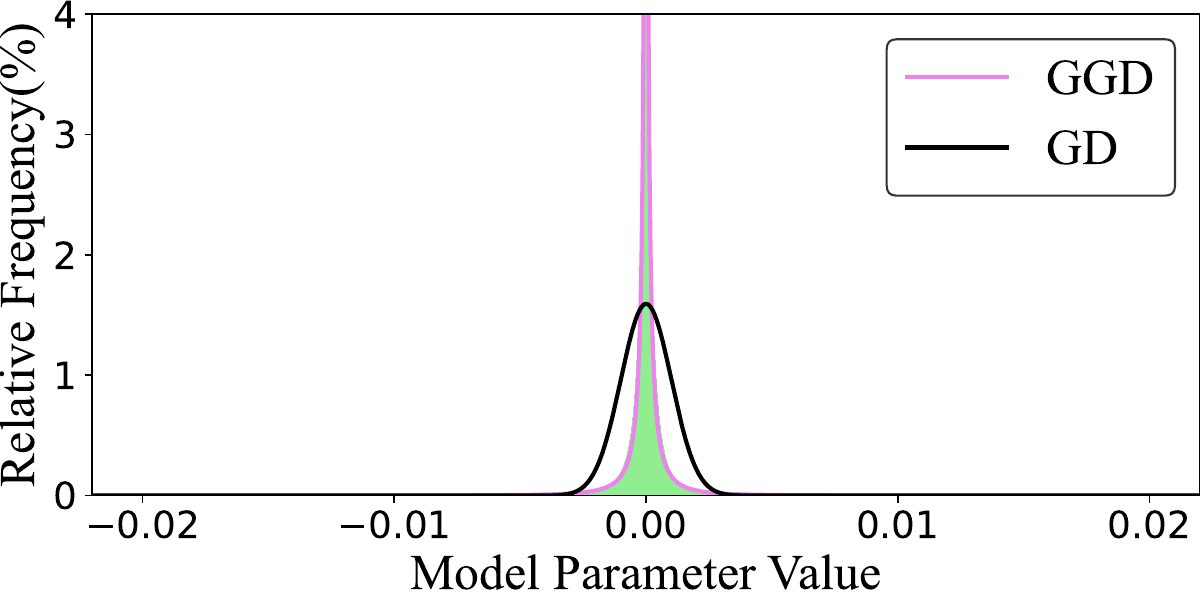}
    \caption{Phi Gradients.}
\end{subfigure}
\begin{subfigure}[b]{0.33\textwidth}
    \centering
    \includegraphics[width=\linewidth]{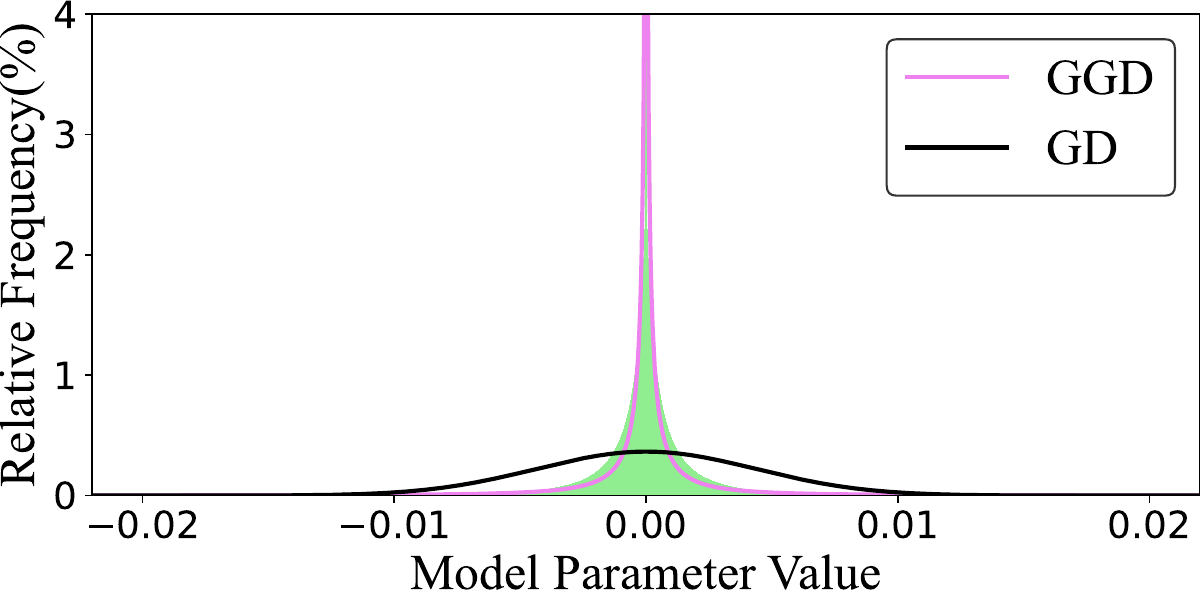}
    \caption{Qwen Gradients.}
\end{subfigure}

\caption{Statistical distributions of weights, activations, and gradients in full-parameter fine-tuning of certain models, fitted with Generalized Gaussian Distribution (GGD) and Gaussian Distribution (GD), respectively. The statistics for weights and gradients are computed over all trainable parameters, while activations are collected from the output of each Transformer block during forward propagation. It can be observed that GGD achieves better fitting performance than GD in all cases.}
\label{fig:dists}
\vskip -0.1in
\end{figure*}

\begin{figure*}[ht]
    \centering
    \includegraphics[width=1\textwidth]{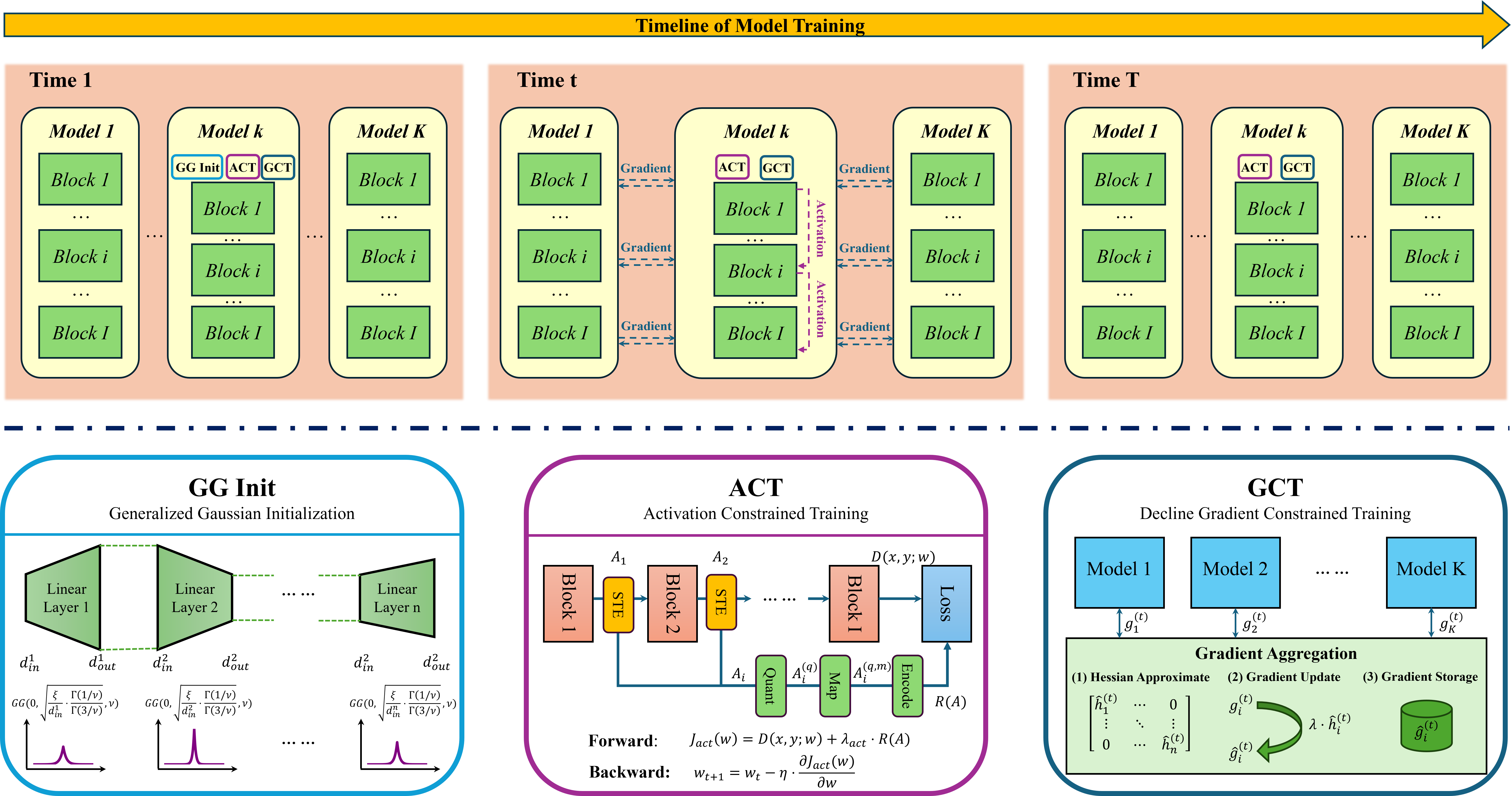}
    \caption{Overview of training process by GG Init, ACT, and GCT. At the initialization phase of training, GG Init is responsible for endowing the model with a low-entropy initial distribution. As training progresses, ACT and GCT are responsible for compressing the gradients generated during forward propagation and the gradients transmitted between models, respectively.}
    \label{fig:sys}

\end{figure*}

\section{Related Work}
\subsection{Training Initialization for LLMs}
Proper initialization is crucial for training LLMs, as it directly affects convergence speed, generalization, and training stability. Classical schemes—including Gaussian, Xavier \citep{glorot2010understanding}, and He initialization \citep{he2015delving}—regulate variance and gradient flow, with Gaussian initialization playing a central role in symmetry breaking. Recent probabilistic analyses revisit initialization statistics \citep{wolinski2025gaussianpreactivationsneuralnetworks, basteri2022quantitative} and explore Gaussian mixtures for improved convergence \citep{shi2024avoidingbarrenplateausgaussian}, motivating more expressive, distribution-aware initialization strategies.

Xavier initialization preserves activation and gradient variance across layers and was later extended to support deeper networks and diverse nonlinearities \citep{he2015delving, kumar2017weight}. He initialization further adapts variance scaling to ReLU activations, mitigating vanishing gradients in rectified networks. Beyond these general schemes, orthogonal initialization stabilizes variance in recurrent models \citep{saxe2014exactsolutionsnonlineardynamics}, IDInit preserves identity mappings in residual networks \citep{pan2025idinituniversalstableinitialization}, and DaWin enables training-free, inference-time adaptation via entropy-driven weight adjustment \citep{oh2025dawin}. Together, these works underscore the central role of principled initialization in stable and efficient training.

\subsection{Activation Compression of Latent Space}

The rapid scaling of neural networks has outpaced growth in VRAM capacity, making distributed training and inference unavoidable for LLMs and exposing activation memory and communication as primary bottlenecks. Prior work mitigates these costs through activation quantization or sparsification, addressing outliers via clipping, scaling, structural transformations, or training-aware parameter learning \citep{pmlr-v202-xiao23c, NEURIPS2024_b5b93943, esser2020learned, choi2018}, and inducing sparsity through ReLU or MoE structures, dynamic activation prediction, or architectural conversion in LLMs \citep{agarap2019deeplearningusingrectified, shazeer2017outrageously, liu2023dejavucontextualsparsity, mirzadeh2023relustrikesbackexploiting, pmlr-v202-nikdan23a}. Low-rank and encoding methods further reveal substantial redundancy in activation spaces \citep{shamshoum2025, wang2021, gist2018}. However, these approaches largely operate post hoc on fixed activation distributions, leaving unexplored the possibility of training models to natively generate low-entropy, well-structured activations that are intrinsically compressible at the source.

\subsection{Gradient Compression in Distributed Training}

In distributed training of large-scale models, communication bandwidth is often far more constrained than computation, making gradient exchange a dominant bottleneck and motivating extensive work on gradient compression. Existing methods largely fall into quantization and sparsification. Quantization reduces communication via low-bit representations, ranging from early 1-bit and ternary schemes \citep{seide20141, wen2017terngrad} to unbiased stochastic quantization and error-feedback methods that preserve convergence \citep{alistarh2017qsgd, wu2018error, karimireddy2019error}, with recent work exploring adaptive, rate–distortion–guided strategies \cite{bernstein2018signsgd, faghri2020adaptive}. Sparsification transmits only salient gradient elements using Top-$k$, sampling, or accumulation-based schemes \cite{aji2017sparse, basu2019qsparse, stich2018sparsified, lin2018deep, song2021communication}, while hybrid approaches jointly combine sparsity and quantization for greater efficiency \citep{jiang2018linear, xie2024jointsq}. Despite their effectiveness, these methods primarily exploit numerical redundancy in pre-existing gradients without shaping the underlying gradient distribution. A natural next step—analogous to activation optimization—is to train models to natively produce low-entropy, compression-friendly gradient distributions, providing superior inputs for downstream communication-efficient training.

\subsection{Optimized LLM Training using Exp-Golomb (EG) Codes and GGDs}

A recent work \citet{pmlr-v267-wu25ai} introduces a training-time compression framework, named BackSlash, that formulates large model optimization as a rate-distortion problem, rather than treating compression as a post hoc step. It further models LLM weights using the GGD and EG codes for entropy coding and produces models that are both sparse and quantization-friendly. It is reported that this approach can reduce parameter storage by up to 90\% while maintaining accuracy across multiple large language model architectures and tasks. Although this work establishes a promising foundation for integrating statistical modeling into the training dynamics of scalable and hardware-efficient models, it leaves many questions unanswered, some of which we are trying to answer in current research.

\section{Generalized Gaussian Prior}
\label{section-3}

BackSlash \citep{pmlr-v267-wu25ai} introduces a training-time compression framework that casts large-model optimization as a rate–distortion problem rather than a post hoc procedure. By modeling LLM weights with generalized Gaussian distributions and applying entropy coding, it produces models that are both sparse and quantization-friendly, achieving up to 90\% parameter storage reduction with negligible accuracy loss across architectures and tasks. However, BackSlash focuses exclusively on weight distributions and does not address how statistical modeling can shape initialization, activations, or gradients during training—leaving substantial efficiency gains untapped and motivating a more comprehensive, end-to-end approach.
\begin{itemize}
    \item \textbf{Entropy Encoding and Communication.} We adopt EG coding, which is well suited for low-shape generalized Gaussian sources \citep{761289}, offering high hardware parallelism and differentiable code length. This enables direct entropy regularization during training and ensures consistency between training-time and communication bitrates.
    \item \textbf{GG-Based Weight Initialization.} We initialize weights from low-entropy generalized Gaussian distributions, yielding models that retain lower entropy and improved compressibility after training (Table~\ref{tab:entropy_shape}).
    \item \textbf{Activation and Gradient Bitrate Control.} Leveraging the strong GG characteristics of activations and gradients, we incorporate EG code length into the loss to enforce bitrate constraints throughout training.
\end{itemize}

\begin{table}[h]
\centering
\caption{GG Entropy of BERT model weights under different shape parameters before and after training.}
\label{tab:entropy_shape}
\begin{tabular}{ccccc}
\toprule
\textbf{Shape} & \textbf{0.1} & \textbf{0.5} & \textbf{1.0} & \textbf{2.0} \\
\midrule
Initial Entropy & 2.62 & 6.79 & 7.22 & 7.38 \\
Trained Entropy & 2.56 & 4.89 & 5.03 & 5.18 \\
\bottomrule
\end{tabular}
\end{table}

\section{GG in LLM Training}

\subsection{GG Initialization}
\label{subsection:GG init}

A GGD can be expressed as
\begin{equation}
\label{eq:ggd}
    f(x;\mu,\beta,\nu)=\frac{\nu}{2\beta\Gamma(1/\nu)}e^{-(\frac{|x-\mu|}{\beta})^{\nu}},
\end{equation}
where
\begin{equation}
    \beta=\sigma\sqrt{\frac{\Gamma(1/\nu)}{\Gamma{(3/\nu)}}},\\
    \; \Gamma(x)=\int_{0}^{\infty} t^{x-1}e^{-t}dt,
\end{equation}
and $\mu$, $\beta$,$\nu$ are the location, scale, and shape parameters, respectively. Obviously, the Gaussian and Laplacian distributions are special cases of the GGD. 

Recent work \citet{pmlr-v267-wu25ai} shows that the weights of various LLMs are well modeled by GGDs. While BackSlash formulates LLM training as a rate–distortion optimization using a GGD prior, it still relies on Gaussian initialization, despite parameters ultimately converging to GGDs. This suggests that directly initializing weights from GGDs may accelerate convergence, improve performance, or both.

Consider a linear layer with input $X^{n\times d_{in}}$, output $Y^{n\times d_{out}}$, and weight matrix $W^{d_{in}\times d_{out}}$. Following standard initialization analysis, we assume $x\in X$, $y \in Y$, and $w \in W$ are i.i.d., zero-mean GGD variables, with $x$ and $w$ independent and sharing common shape and scale parameters. Under these assumptions, the variance of a GGD is given by
\begin{equation}
\label{eq:ggvar}
    \mathbb{D}[x]=\int_{-\infty}^{x}t^2f(t;0,\beta,\nu)dt=
\beta^2\frac{\Gamma(3/\nu)}{\Gamma(1/\nu)}.
\end{equation}
It can be seen that the variance $\mathbb{D}[x]$ is determined solely by $\nu$ and $\beta$. Moreover, during forward propagation, we have $\nu_x = \nu_y$ and $\beta_x = \beta_y$, so $\mathbb{D}[x]=\mathbb{D}[y]$. and for the $k$-th dimension $y_{i,k} \in Y$ of any sample $y_i$, 
$y_{i,k} = \sum_{j=1}^{d_{\text{in}}} w_{j,k} x_{i,j}$, so $\mathbb{D}[y_{i,k}] = \mathbb{D}\left[ \sum_{j=1}^{d_{\text{in}}} w_{j,k} x_{i,j} \right]$. Since $x$, $y$, and $w$ are i.i.d., and $w$ is independent of $x$, it follows that $\mathbb{D}[y] = d_{\text{in}} \cdot \mathbb{D}[w] \cdot \mathbb{D}[x]$, and therefore
\begin{equation}
\label{eq:gg1}
    \beta_w=\sqrt{\frac{1}{d_{in}}\cdot\frac{\Gamma(1/\nu_w)}{\Gamma(3/\nu_w)}}.
\end{equation}
Consider the effect of different activation functions have on the variance of neuron output distributions, we introduce a correction coefficient $\xi$ to adjust the variance, which is set to 1 for the case without an activation function or with a Sigmoid, $\xi = 2$ for ReLU, and $\xi = \frac{2}{1 + k^2}$ for Leaky-ReLU, where $k$ is the slope of the negative half-axis \citep{he2015delving}. And therefore, GG initialization for linear layers in neural networks
\begin{equation}
    W^{d_{in}\times d_{out}}\sim GG(0,\sqrt{\frac{\xi}{d_{in}}\cdot\frac{\Gamma(1/\nu)}{\Gamma(3/\nu)}},\nu)
\end{equation}
The shape parameter $\nu$ acts as a hyperparameter in the GG initialization. When $\nu=2$, GG initialization degenerates to He initialization. 

\subsection{Activation Constrained Training (ACT)}

Based on information rate-distortion optimization theory \citep{5311476}, the model optimization objective with activation value rate constraint $J_{\text{act}}(w)$ can be expressed as:
\begin{equation}
    \mathcal{J}_{\text{act}}(w)=\mathcal{D}(X,Y;w)+\lambda_{\text{act}}\cdot \sum_{i=1}^{n_{c}}\mathcal{R}(A_{p_i})
\end{equation}
where $\mathcal{D}(\cdot)$ is the distortion function (e.g., mean squared error, cross-entropy, etc.). $\mathcal{R}(\cdot)$ is the rate function of the activation values $\{A_{p_1}, A_{p_2}, \cdots, A_{p_{n_c}}\}$ output by various modules across different devices. Here, $P=\{p_1, p_2, \cdots, p_{n_c}\}\subseteq [L]$ is the subset of activation layer indices determined for a given model parallel partitioning scheme, and $L$ denotes the total number of layers in the model. $\lambda_{\text{act}}$ is the Lagrange multiplier, used to achieve a trade-off between distortion and rate.

As discussed in Section~\ref{section-3}, model states follow generalized Gaussian statistics, for which Exponential-Golomb (EG) coding \citep{eg1977code,761289} is well suited, enabling shared-codebook encoding, high hardware parallelism, and differentiable code length. This allows the average EG code length to serve directly as the rate function $\mathcal{R}(\cdot)$, aligning optimization with true communication cost. Prior to encoding, floating-point activations are linearly quantized to integers with step size $2^{-n}$. For the $n_p$-dimensional activation vector $A_p$, we apply a straight-through estimator (STE) \citep{liu2022nonuniform}, $\text{STE}(x)=x-\text{sg}(x-\lfloor x\rfloor)$, to decouple forward quantization from backward gradients. The quantization operation is given by $A_p^{(q)}=\text{Quant}(A_p)=\text{STE}(2^n\cdot A_p)$. The quantized activation values $A_p^{(q)}$ need to be mapped to non-negative elements to facilitate subsequent unsigned integer encoding:
\begin{equation}
A_{p}^{(q,m)}=\text{Map}(A_p^{(q)})=
\begin{cases}
-2A_p^{(q)} & A_p^{(q)}<0 \\
0 & A_p^{(q)}=0 \\
2A_p^{(q)}-1 & A_p^{(q)}>0
\end{cases}
\end{equation}
The unsigned elements $A_p^{(q,m)}=\text{Map}(\text{Quant}(A_p))$ can then be directly encoded into code lengths based on element values using $k$-th order EG encoding. Therefore, the rate function $\mathcal{R}(\cdot)$ based on EG code can be derived as:
\begin{equation}
\mathcal{R}(A_p)=\frac{2}{n_p}\cdot\sum_{j=1}^{n_p}\cdot\text{STE}\left( \log_2\left(A_{p,j}^{(q,m)}\right)+2^k\right)-k+1
\end{equation}

Then the training process of the ACT algorithm can be described as follows.
\renewcommand{\thealgorithm}{1}
\begin{algorithm}[htbp]
    \caption{Activation Constrained Training (ACT)}
    \label{alg:act}
    \begin{algorithmic}[1]
        \STATE \textbf{Require:} Distortion function $\mathcal{D}$, model weights $w$, learning rate $\eta$, Lagrange multiplier $\lambda_{act}$, quantization step $2^{-n}$, EG order $k$, decay factor $\alpha$.
        \STATE \textbf{Initialize:} $w^{(0)}$, $t \leftarrow 0$, $n_c$ for number of model shards.
        \FOR{each epoch $\tau = 0, 1, 2, \dots$}
            \FOR{each batch $(x_b, y_b)$}
                \STATE Forward and calculate distortion $\mathcal{D}(x_b,y_b;w)$ and activations $\{p_1, p_2, \cdots, p_{n_c}\}$ with quantization.
                \STATE Calculate activation rate $\mathcal{R}(A)\leftarrow\sum_{i=1}^{n_c}\frac{2}{n_{p_i}}\cdot\sum_{j=1}^{n_{p_i}}\cdot\text{STE}\left( \log_2\left(\text{Map}(\text{STE}(2^n \cdot A_{p_i,j}))\right)+2^k\right)-k+1$.
                \STATE Compute the RD Cost $\mathcal{J}_{act}(w)\leftarrow \mathcal{D}(x_b,y_b;w)+\lambda_{act}\cdot\mathcal{R}(A)$.
                \STATE Back propogation and update the weights $w\leftarrow w-\eta\cdot\frac{\partial \mathcal{J}_{act}(w)}{\partial w}$.
            \ENDFOR
        \ENDFOR
        \STATE \textbf{Until} convergence or max iterations.
    \end{algorithmic}
\end{algorithm}
ACT reduces activation communication and memory cost while enabling information-theoretic analysis of neural representations. By treating activations as latent codes, their rate characterizes the model’s information bottleneck and representation efficiency. Constraining parameters to shape activation distributions further provides a principled means to identify influential neurons and their hierarchical contributions to information flow.

\subsection{Gradient Constrained Training (GCT)}
\begin{figure*}[h!]
    \centering
    \begin{subfigure}{0.32\textwidth}
        \centering
        \includegraphics[width=\linewidth]{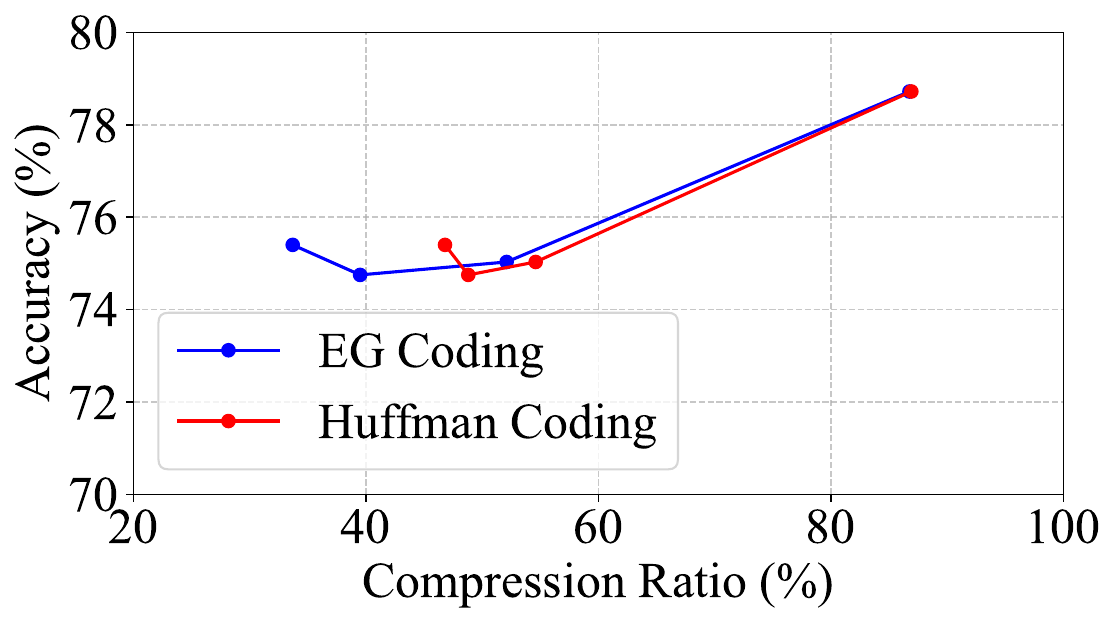}
        \caption{Trade-off of GG Initialization.}
    \end{subfigure}
    \hfill
    \begin{subfigure}{0.32\textwidth}
        \centering
        \includegraphics[width=\linewidth]{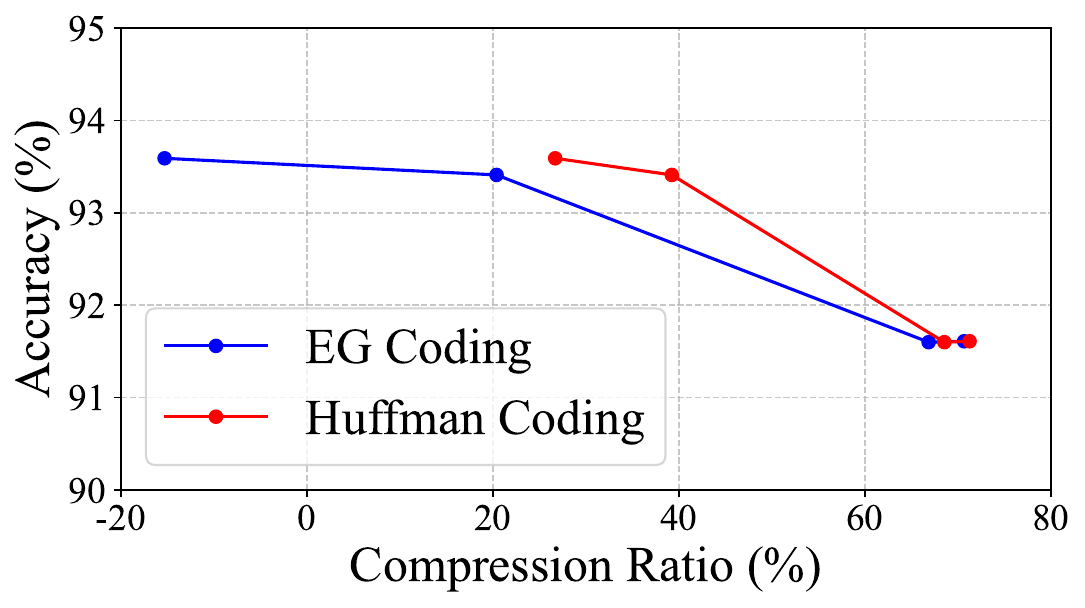}
        \caption{Trade-off of ACT.}
    \end{subfigure}
    \hfill
    \begin{subfigure}{0.32\textwidth}
        \includegraphics[width=\linewidth]{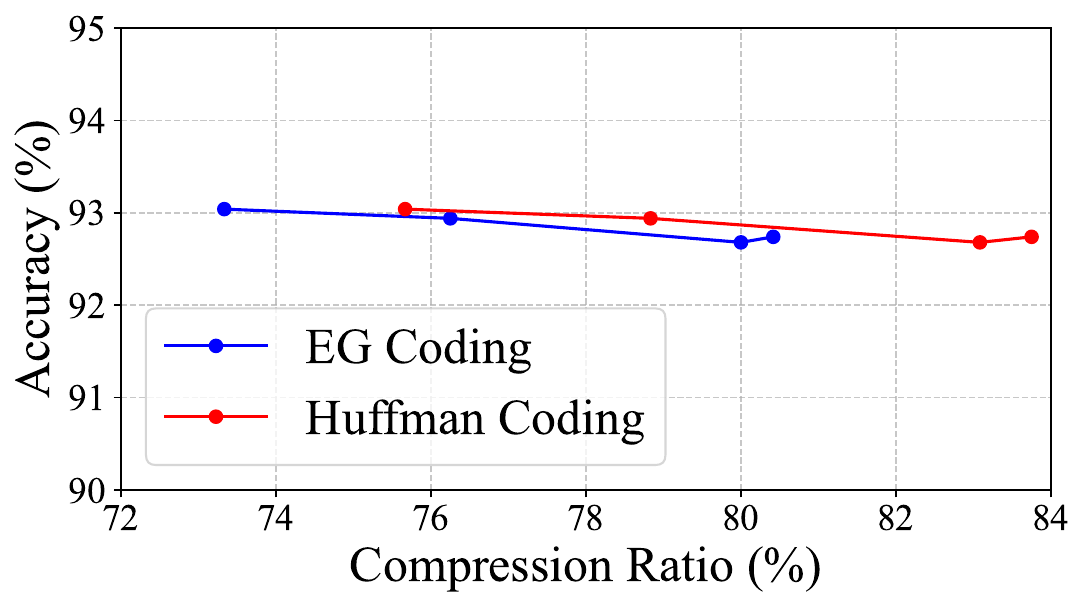}
        \caption{Trade-off of GCT.}
    \end{subfigure}
    
    \caption{The trade-off between model accuracy and compression rate of GG Initialization, ACT, and GCT under various hyper-parameters where the shape parameter varies in GGI, and ACT and GCT utilize different Lagrange multipliers.}
    \vskip -0.1in
    \label{fig:trade-off}
\end{figure*}

\begin{figure*}[ht!]
    \centering
        \begin{subfigure}{0.49\textwidth}
        \centering
        \includegraphics[width=\linewidth]{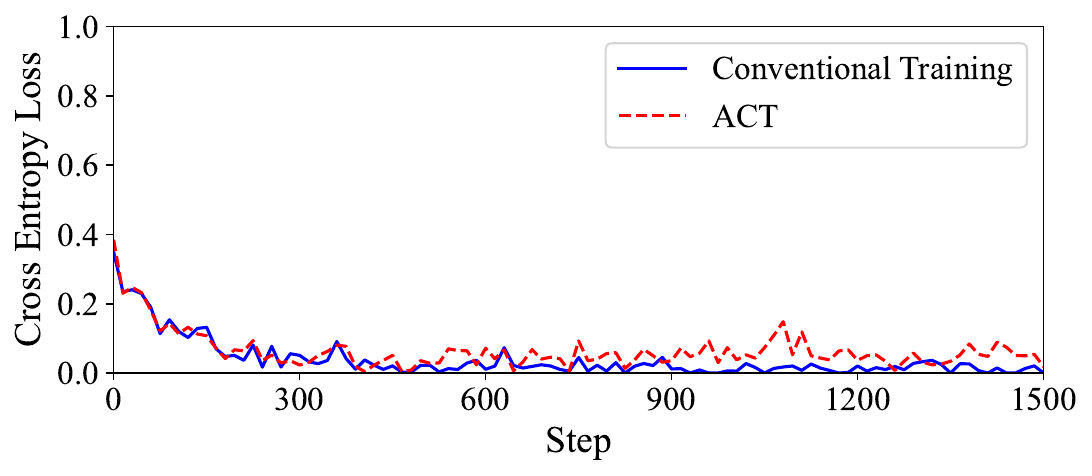}
        \caption{Evolution of training loss in ACT.}
        \label{fig:act-loss}
    \end{subfigure}
    \hfill
    \begin{subfigure}{0.49\textwidth}
        \centering
        \includegraphics[width=\linewidth]{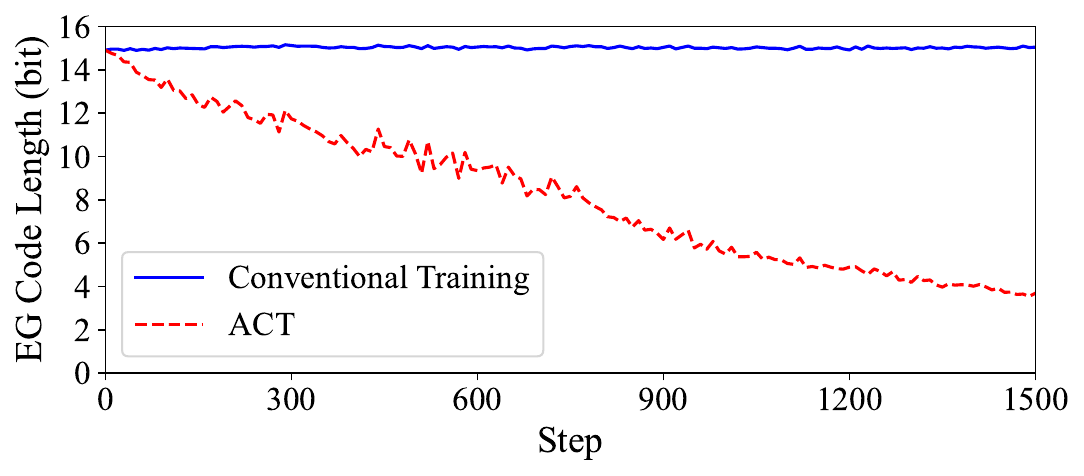}

        \caption{Evolution of EG code length in ACT.}
        \label{fig:act-eg}
    \end{subfigure}
    \hfill
    \begin{subfigure}{0.49\textwidth}
        \centering
        \includegraphics[width=\linewidth]{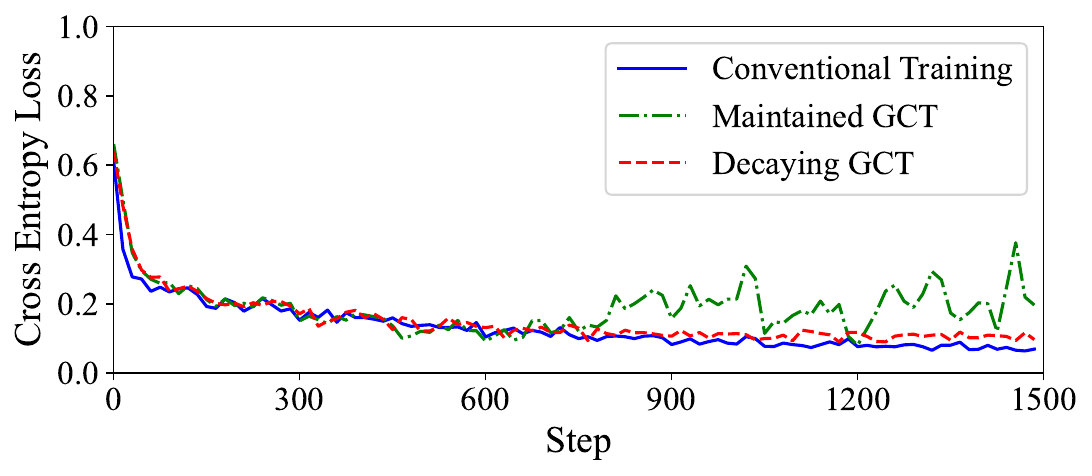}
        \caption{Evolution of training loss in GCT.}
        \label{fig:gct-loss}
    \end{subfigure}
    \hfill
    \begin{subfigure}{0.49\textwidth}
        \centering
        \includegraphics[width=\linewidth]{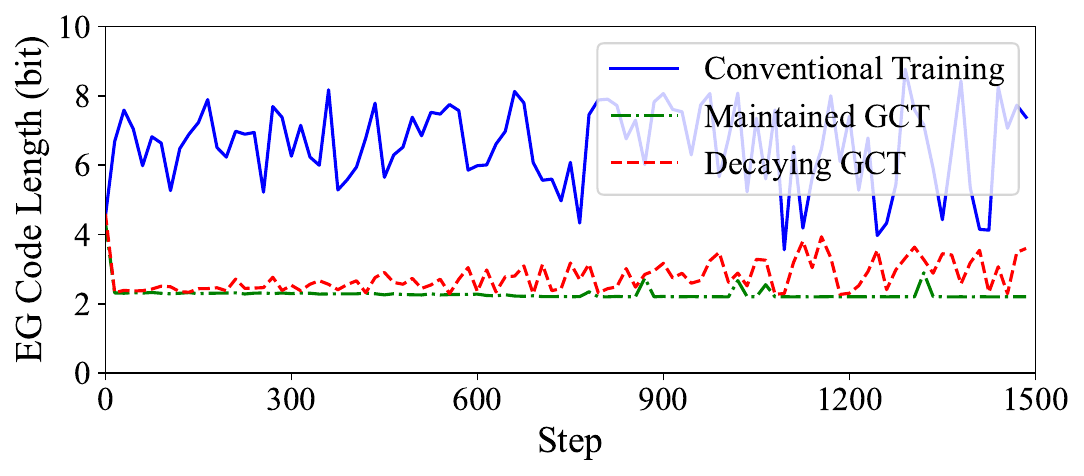}
        \caption{Evolution of EG code length in GCT.}
        \label{fig:gct-eg}
    \end{subfigure}
    
    \caption{The variation of loss and average EG code length for ACT, GCT, compared to conventional training.}
    \vskip -0.2 in
    \label{fig:training-process}
\end{figure*}

Similarly, assuming the gradient $G=\frac{\partial \mathcal{D}(X,Y;w)}{\partial w}$, the model optimization objective with gradient rate constraint $\mathcal{J}_{\text{grad}}(w)$ can be expressed as:
\begin{equation}
    \label{eq:grad-1}
    \mathcal{J}_{grad}(w)=\mathcal{D}(X,Y;w)+\lambda_{grad}\cdot \mathcal{R}(G)
\end{equation}
where $X$, $Y$, and $w$ represent the input, output, and model weights, respectively. And $\mathcal{D}(\cdot)$, $\mathcal{R}(\cdot)$, $\lambda_{grad}$ denote the distortion, rate, and Lagrange multiplier, respectively. $\mathcal{R(\cdot)}$ uses EG code length for implementation. Meanwhile, to increase the parallelism of gradient constraints during weight updates, we directly present the update formula incorporating the gradient-constrained loss:
\begin{equation}
\label{eq-3}
    w_i^{(t+1)}=w_i^{(t)}-\eta\cdot\bigg(\frac{\partial \mathcal{D}^{(t)}}{\partial w_i^{(t)}}+\lambda_{grad}\cdot\frac{\partial \mathcal{R}^{(t)}}{\partial w_i^{(t)}}\bigg)
\end{equation}
where $\frac{\partial \mathcal{R}^{(t)}}{\partial w_i^{(t)}}=\frac{\partial \mathcal{R}^{(t)}}{\partial G^{(t)}}\cdot \sum_{j=1}^{n}\frac{\partial^2 \mathcal{D}^{(t)}}{\partial w_j^{(t)} \partial w_i^{(t)}}$, and $\frac{\partial\mathcal{R}^{(t)}}{\partial G^{(t)}}=\frac{2^{n+2}}{n\ln 2}\cdot \frac{\text{Sign}(G^{(t)})}{\text{Map}(\text{Quant}(G^{(t)}))}$.

Since $\partial^2 \mathcal{D}/\partial w_j \partial w_i$ couples parameters across layers, second-order propagation can only proceed after full backpropagation of $\partial \mathcal{D}/\partial w_i$. This would require computing and storing the full Hessian, imposing prohibitive computational and memory costs.
A common approximation \citep{pmlr-v28-sutskever13,bottou2018optimization,neal1998view} retains only the diagonal of the Hessian, setting $\partial^2 \mathcal{D}/\partial w_j \partial w_i=0$ for $j\neq i$. Under this assumption, Equation~\ref{eq-3} simplifies to:
\begin{equation}
\label{eq-4}
    w_i^{(t+1)}=w_i^{(t)}-\eta\bigg(\frac{\partial \mathcal{D}^{(t)}}{\partial w_i^{(t)}}+\lambda_{grad}\frac{\partial \mathcal{R}^{(t)}}{\partial G^{(t)}}\frac{\partial^2 \mathcal{D}^{(t)}}{\partial w_i^{(t)2}}\bigg)
\end{equation}
Exact second-order derivatives require an extra backward pass and are infeasible for billion-parameter LLMs, so we adopt an efficient approximation. Let $g_i^{(t)}=\partial \mathcal{D}^{(t)}/\partial w_i^{(t)}$ and $h_i^{(t)}=\partial^2 \mathcal{D}^{(t)}/\partial w_i^{(t)2}$; in GCT the effective gradient is $\hat{g}_i^{(t)}=g_i^{(t)}+\lambda h_i^{(t)}$. With the small learning rates used in LLMs, $\delta w=w_i^{(t)}-w_i^{(t-1)}=\eta\cdot\hat g_i^{(t-1)}$ is small, allowing $h_i^{(t)}$ to be approximated by $\hat{h}_i^{(t)}$ as follows:
\begin{equation}
\label{eq-5}
    \hat h_i^{(t)}=\frac{g_i^{(t)}-\hat g_i^{(t-1)}}{w_i^{(t)}-w_i^{(t-1)}}=\frac{1}{\eta}\cdot\left(1-\frac{ g_i^{(t)}}{\hat g_i^{(t-1)}}\right)
\end{equation}
To prevent instability when $\hat g_i^{(t-1)}$ becomes too small, we introduce a small constant $\epsilon$ for gradient clipping. Equation~\ref{eq-5} then becomes:
\begin{equation}
    \hat h_i^{(t)}=-\frac{1}{\eta}\cdot\frac{g_i^{(t)}}{\hat g_i^{(t-1)}+\epsilon}+\frac{1}{\eta}.
\end{equation}
This approach requires storing only the previous-step equivalent gradients to approximate second-order terms. Under this approximation, the weight update in Equation~(4) simplifies to:
\begin{equation}
    w_i^{(t+1)}=w_i^{(t)}-\eta\cdot g_i^{(t)}-\eta\cdot\lambda_{grad}\cdot\frac{\partial \mathcal{R}^{(t)}}{\partial g_i^{(t)}}\cdot \hat h_i^{(t)}.
\end{equation}
A fixed Lagrange multiplier late in training can induce loss oscillations and impede convergence. We therefore apply a decay factor $0<\alpha<1$, updating $\lambda$ after epoch $\tau$ as $\lambda_{\tau+1} = \alpha \cdot \lambda_\tau$.
And the GCT algorithm can be summarized as follows:
\renewcommand{\thealgorithm}{2}
\begin{algorithm}
    \caption{Decaying Gradient Constrained Training Algorithm (GCT)}
    \begin{algorithmic}[1]
        \STATE \textbf{Require:} Distortion function $\mathcal{D}$, weights $w$, training step $t=1$, learning rate $\eta$, Lagrange multiplier $\lambda$, smoothing coefficient $\epsilon$, decaying coefficient $\alpha$.
        \FOR{each epoch $\tau$}
            \FOR{each batch $(x_b, y_b)$}
                \STATE Forward and back propagation to compute the gradients: $g_i^{(t)}\leftarrow\partial\mathcal{D}^{(t)}/\partial w_i^{(t)}$.
                \STATE Compute approximate second-order derivatives: $\hat h_i^{(t)}\leftarrow(-\frac{1}{\eta}\cdot\frac{g_i^{(t)}}{\hat g_i^{(t-1)}+\epsilon}+\frac{1}{\eta})\cdot \mathbf{1}_{t \ge 2}$.
                \STATE Compute and store the equivalent gradient: $\hat{g}_{i}^{(t)}\leftarrow g_i^{(t)}+\lambda_{\tau}\cdot \partial \mathcal{R}^{(t)}/\partial g_i^{(t)} \cdot \hat{h}_i^{(t)}$.
                \STATE Weight update: $w_i^{(t+1)}\leftarrow w_i^{(t)}-\eta\cdot \hat{g}_{i}^{(t)}$, $t\leftarrow t+1$.
            \ENDFOR
            \STATE Constraint decay: $\lambda_{\tau+1} = \alpha \cdot \lambda_\tau$.
        \ENDFOR
        \STATE \textbf{Until} convergence or max iterations.
    \end{algorithmic}
\end{algorithm}

\begin{table*}[ht]
\caption{Compression ratio of conventional training and BackSlash using GG initialization with different shape parameters.}
\label{table:init-shape}
\begin{center}
\begin{small}
\begin{tabular}{ccccc|ccc}
\toprule
\textbf{Shape} & \textbf{Method} & \textbf{FL (bits)} & \textbf{EG (bits)} & \textbf{HM (bits)} & \boldmath{$C_{EG}$} & \boldmath{$C_{HM}$} & \textbf{Accuracy}\\
\midrule

\multirow{2}{*}{0.1} & Conventional & 12.00 & 1.59 & 1.57 & 87\% & 87\% & 78.72\% \\
 & BackSlash & 12.00 & 1.10 & 1.10 & \textbf{91\%} & \textbf{91\%} & \textbf{83.20\%} \\
\midrule

\multirow{2}{*}{0.5} & Conventional & 10.00 & 4.97 & 4.54 & 50\% & 55\% & 75.03\% \\
& BackSlash & 10.00 & 1.99 & 1.90 & \textbf{80\%} & \textbf{81\%} & \textbf{81.68\%} \\
\midrule

\multirow{2}{*}{1.0} & Conventional & 10.00 & 6.05 & 5.12 & 39\% & 49\% & 74.75\% \\
& BackSlash & 10.00 & 2.18 & 2.06 & \textbf{78\%} & \textbf{79\%} & \textbf{81.87\%} \\
\midrule

\multirow{2}{*}{2.0} & Conventional & 10.00 & 6.63 & 5.32 & 34\% & 47\% & 75.40\% \\
 & BackSlash & 10.00 & 2.27 & 2.12 & \textbf{77\%} & \textbf{79\%} & \textbf{78.18\%} \\
\bottomrule
\end{tabular}
\end{small}
\end{center}
\vskip -0.1in
\end{table*}

\begin{table*}[ht]
\caption{Compression ratio of conventional training by GG Init using DeepSeek models of different scales.}
\label{table:deepseek-scale}
\begin{center}
\begin{small}
\begin{tabular}{cccc|cccc}
\toprule
\textbf{Model} & \textbf{FL (bits)} & \textbf{EG (bits)} & \textbf{HM (bits)} & \boldmath{$C_{EG}$} & \boldmath{$C_{HM}$} & \textbf{Train Accuracy} & \textbf{Test Accuracy}\\
\midrule

DeepSeek-1.5B & 11.00 & 4.77 & 3.95 & 57\% & 64\% & 99.49\% & 23.51\% \\
DeepSeek-7B   & 11.00 & 4.29 & 3.50 & 61\% & 68\% & 99.81\% & 25.30\% \\
DeepSeek-8B   & 11.00 & 4.12 & 3.34 & 63\% & 70\% & 99.93\% & 23.18\% \\
DeepSeek-14B  & 11.00 & 4.10 & 3.38 & 63\% & 69\% & 99.78\% & 21.89\% \\

\bottomrule
\end{tabular}
\end{small}
\end{center}
\vskip -0.1in
\end{table*}

\section{Experiment}
\subsection{Settings}
We evaluate GG Init, ACT, and GCT through extensive experiments across diverse models and tasks, analyzing training dynamics, generalization, and deployment efficiency. Experiments span BERT, GPT, LLaMA, and DeepSeek models at multiple scales, with text classification on IMDB and Spam and text generation on WMT and SQuAD. Performance is measured by accuracy or next-token accuracy. To quantify compression effects, we follow \citep{pmlr-v267-wu25ai} and report average code lengths using Exponential-Golomb (EG) and Huffman (HM) entropy coding, benchmarked against Fixed-Length (FL) coding; $C_{EG}$ and $C_{HM}$ denote compression ratios relative to FL.

\subsection{Trade-off}

\begin{table*}[ht]
\caption{Generalization performance of GG Init, ACT, and GCT. The unconstrained baselines (He initialization for GG Init; conventional training for ACT/GCT) are compared against their constrained counterparts (GG initialization with shape 0.1; constrained training).}
\label{table:tasks}
\begin{center}
\small
\begin{tabular}{cccc|ccc|ccc}
\toprule
\textbf{Model} & \textbf{Dataset} & \textbf{Method} & \textbf{Strategy} & \textbf{EG} & \textbf{HM} & \textbf{FL} & \boldmath{$C_{EG}$} & \boldmath{$C_{HM}$} & \textbf{Accuracy} \\
\midrule

\multirow{6}{*}{BERT-110M} & \multirow{6}{*}{IMDB} 
& \multirow{2}{*}{GG Init} & Unconstrained & 6.63 & 5.32 & 10.00 & 33.70\% & 46.80\% & 75.40\% \\
& & & Constrained & 1.59 & 1.57 & 12.00 & \textbf{86.75\%} & \textbf{86.92\%} & \textbf{78.72\%} \\
\cmidrule(lr){3-10}
& & \multirow{2}{*}{ACT} & Unconstrained & 14.99 & 9.53 & 13.00 & -15.31\% & 26.69\% & \textbf{93.59\%} \\
& & & Constrained & 4.31 & 4.09 & 13.00 & \textbf{66.85\%} & \textbf{68.54\%} & 91.62\% \\
\cmidrule(lr){3-10}
& & \multirow{2}{*}{GCT} & Unconstrained & 7.58 & 6.18 & 11.00 & 31.09\% & 43.82\% & 92.94\% \\
& & & Constrained & 3.43 & 3.16 & 12.00 & \textbf{71.42\%} & \textbf{73.67\%} & \textbf{92.97\%} \\

\midrule

\multirow{6}{*}{GPT-774M} & \multirow{6}{*}{Spam} 
& \multirow{2}{*}{GG Init} & Unconstrained & 5.83 & 4.72 & 9.00 & 35.22\% & 47.56\% & \textbf{99.25\%} \\
& & & Constrained & 2.16 & 2.12 & 11.00 & \textbf{80.36\%} & \textbf{80.73\%} & 99.10\% \\
\cmidrule(lr){3-10}
& & \multirow{2}{*}{ACT} & Unconstrained & 18.22 & 11.27 & 15.00 & -21.47\% & 24.87\% & \textbf{99.15\%} \\
& & & Constrained & 3.07 & 2.64 & 12.00 & \textbf{74.42\%} & \textbf{78.00\%} & 97.60\% \\
\cmidrule(lr){3-10}
& & \multirow{2}{*}{GCT} & Unconstrained & 6.93 & 5.84 & 12.00 & 42.25\% & 51.33\% & \textbf{98.05\%} \\
& & & Constrained & 2.97 & 2.60 & 12.00 & \textbf{75.25\%} & \textbf{78.33\%} & 95.15\% \\

\midrule

\multirow{6}{*}{LLaMA-1B} & \multirow{6}{*}{WMT} 
& \multirow{2}{*}{GG Init} & Unconstrained & 5.63 & 4.47 & 9.00 & 37.44\% & 50.33\% & 46.13\% \\
& & & Constrained & 4.54 & 3.55 & 11.00 & \textbf{58.73\%} & \textbf{67.73\%} & \textbf{51.99\%} \\
\cmidrule(lr){3-10}
& & \multirow{2}{*}{ACT} & Unconstrained & 10.27 & 7.22 & 11.00 & 6.64\% & 34.36\% & \textbf{70.79\%} \\
& & & Constrained & 4.37 & 3.89 & 11.00 & \textbf{60.27\%} & \textbf{64.64\%} & 70.67\% \\
\cmidrule(lr){3-10}
& & \multirow{2}{*}{GCT} & Unconstrained & 6.72 & 5.58 & 12.00 & 44.00\% & 53.50\% & \textbf{48.44\%} \\
& & & Constrained & 2.75 & 2.32 & 12.00 & \textbf{77.08\%} & \textbf{80.67\%} & 45.35\% \\

\midrule

\multirow{6}{*}{DeepSeek-7B} & \multirow{6}{*}{SQuAD} 
& \multirow{2}{*}{GG Init} & Unconstrained & 5.26 & 4.16 & 10.00 & 47.40\% & 58.40\% & 25.18\% \\
& & & Constrained & 4.29 & 3.50 & 11.00 & \textbf{61.00\%} & \textbf{68.18\%} & \textbf{25.30\%} \\
\cmidrule(lr){3-10}
& & \multirow{2}{*}{ACT} & Unconstrained & 17.46 & 10.74 & 12.00 & -45.50\% & 10.50\% & \textbf{70.79\%} \\
& & & Constrained & 7.72 & 5.95 & 12.00 & \textbf{35.67\%} & \textbf{50.42\%} & 70.67\% \\
\cmidrule(lr){3-10}
& & \multirow{2}{*}{GCT} & Unconstrained & 4.66 & 4.24 & 12.00 & 61.17\% & 64.67\% & 45.10\% \\
& & & Constrained & 2.47 & 1.94 & 12.00 & \textbf{79.42\%} & \textbf{83.83\%} & \textbf{45.31\%} \\

\bottomrule
\end{tabular}
\end{center}
\vskip -0.2in
\end{table*}

We analyze the bitrate–performance trade-off to assess robustness under compression and guide hyperparameter selection for resource-constrained distributed training. Using BERT on IMDB, we vary the GG Init shape parameter and the ACT/GCT Lagrange multipliers to obtain performance–bitrate curves. As shown in Fig.~\ref{fig:trade-off}, GG Init exhibits a synchronous increase in accuracy and bitrate under both coding schemes, suggesting that lower shapes ease optimization and improve generalization. In contrast, ACT and GCT show an inverse trade-off, with accuracy remaining largely stable as bitrate decreases. Despite mild fluctuations, both methods achieve substantial compression with minimal performance loss, demonstrating that constrained training can significantly reduce communication cost while preserving accuracy in practical distributed settings.

\subsection{Training Performance}

As ACT and GCT regulate dynamically generated activations and gradients, tracking loss and bitrate during training reveals their underlying behaviors. Using BERT on IMDB, we compare ACT and GCT to conventional training by monitoring loss and EG code length.
For ACT, Fig.~\ref{fig:act-loss} shows loss curves nearly identical to conventional training, indicating unaffected convergence. Conversely, Fig.~\ref{fig:act-eg} shows ACT progressively reduces activation bitrate, while conventional training remains high, reflecting ACT’s indirect but effective shaping of activation distributions.
For GCT, Fig.~\ref{fig:gct-loss} demonstrates that decaying GCT ($\alpha=0.8$) closely follows conventional training, whereas non-decay GCT ($\alpha=1.0$) suffers late-stage oscillations. Fig.~\ref{fig:gct-eg} further shows that decaying GCT yields a lower and smoother gradient bitrate than conventional training, which exhibits large fluctuations. These results show that ACT and GCT substantially reduce communication cost with stable convergence, highlighting the importance of distribution-aware gradient optimization.

\subsection{Ablation Analysis of GG Init}
GG initialization modifies only parameter initialization of LLMs without extra computational cost, requiring only the choice of shape parameters. Using BERT on IMDB, we evaluate shapes $\{0.1,0.5,1.0,2.0\}$ under conventional and BackSlash training in Table~\ref{table:init-shape}. Smaller shapes yield higher compression ratios, while BackSlash outperforms conventional training in both compression and accuracy for all settings. Gaussian initialization (shape $=2$) performs worst, and low-shape GG initialization ($0.1$) strongly constrains post-training code length, highlighting the importance of distribution-aware initialization.
We further apply GG initialization ($0.1$) to DeepSeek models (1.5B–14B) trained on SQuAD (Table~\ref{table:deepseek-scale}). As model size increases, average code length decreases, reflecting lower information density per parameter. Although training raises code length from its low initial value, much of the added information appears noisy when data is limited, indicating that scaling model size does not proportionally increase useful information absorption.

\subsection{Generalization}

We evaluate the generalization of GGI, ACT, and GCT across multiple architectures and tasks. ACT targets activation compression at inference and reports average activation code length on the test set, while GCT optimizes gradients during training on the training set.
As shown in Table~\ref{table:tasks}, GGI consistently achieves higher compression without sacrificing accuracy. Generative tasks exhibit higher final weight code lengths than classification tasks, reflecting their greater modeling complexity. ACT and GCT substantially compress activations and gradients with minimal accuracy loss: ACT shows slightly higher code lengths on generative tasks due to its dependence on learned weight distributions, whereas GCT maintains similar code lengths across tasks, indicating strong task and architecture independence.
Notably, ACT produces significantly lower EG code lengths than FL, while conventional training often exceeds FL due to activation outliers. This confirms that ACT effectively reshapes activation distributions and suppresses outliers. Overall, the results demonstrate that GGI, ACT, and GCT generalize robustly across models and tasks, from classification to generation.

\section{Conclusion}
We present a unified, statistically grounded framework for LLM optimization that leverages generalized Gaussian priors across initialization, training, and communication. By modeling weights, activations, and gradients as heavy-tailed GG distributions, our approach improves convergence, compressibility, and communication efficiency. The framework integrates GG-based initialization, Activation-Constrained Training, and Gradient-Constrained Training, transforming compression from a post hoc step into an intrinsic learning principle. Extensive experiments show consistent gains in compression and efficiency with minimal accuracy loss. This work highlights the value of rate-aware, statistically principled model design and opens avenues for extending GG-informed optimization to multimodal, federated, and energy-efficient AI systems.

\newpage
\section*{Impact Statement}
This paper introduces a framework that redefines LLM optimization by treating efficiency as an intrinsic learning principle rather than a post-processing step. Leveraging generalized Gaussian priors throughout the entire model lifecycle, it enables the development of LLMs that are inherently compressed and high-performing. This work significantly reduces the computational and environmental footprint of training models, while accelerating their deployment across resource-constrained hardware, federated systems, and energy-efficient AI applications.

\bibliography{icml2026}
\bibliographystyle{icml2026}

\end{document}